\documentclass[sigconf,nonacm]{acmart}
\AtBeginDocument{%
  }

\usepackage{listings}
\usepackage{courier}
\usepackage{xcolor}
\usepackage[table]{xcolor}
\usepackage{tabularx}
\usepackage{booktabs}
\usepackage{array}
\usepackage{colortbl}
\usepackage{multirow}
\usepackage{makecell}
\usepackage{caption}
\usepackage{pifont}

\newcommand{\cmark}{\ding{51}}
\newcommand{\xmark}{\ding{55}}

\newcolumntype{R}[1]{>{\hsize=#1\hsize\raggedright\arraybackslash}X}
\newcolumntype{C}[1]{>{\centering\arraybackslash}p{#1}}

\begin{document}
\emergencystretch 7em

\title{An Ontology for Unified Modeling of Tasks, Actions, Environments, and Capabilities in Personal Service Robotics}

\author{Margherita Martorana}
\email{m.martorana@vu.nl}
\orcid{0000-0001-8004-0464}
\affiliation{%
  \institution{Vrije Universiteit}
  \city{Amsterdam}
  \country{The Netherlands}
}

\author{Francesca Urgese}
\orcid{0009-0009-0516-1800}
\affiliation{%
  \institution{Vrije Universiteit}
  \city{Amsterdam}
  \country{The Netherlands}
}

\author{Ilaria Tiddi}
\orcid{0000-0001-7116-9338}
\affiliation{%
  \institution{Vrije Universiteit}
  \city{Amsterdam}
  \country{The Netherlands}
}

\author{Stefan Schlobach}
\orcid{0000-0002-3282-1597}
\affiliation{%
  \institution{Vrije Universiteit}
  \city{Amsterdam}
  \country{The Netherlands}
}

\renewcommand{\shortauthors}{Martorana et al.}

\begin{abstract}
Personal service robots are increasingly deployed to support daily living in domestic environments, particularly for elderly and individuals requiring assistance. Operating effectively in these settings involves not only physical interaction but also the ability to interpret dynamic environments, understand tasks, and choose appropriate actions based on context. This process requires integrating both hardware components (e.g. sensors, actuators) and software systems capable of reasoning about tasks, environments, and the robot's own capabilities. Framework such as the Robot Operating System (ROS) provide open-source libraries and tools that help connect low-level hardware data with higher-level functionalities. However, real-world deployments remain tightly coupled to specific hardware and software platforms. As a result, solutions are often isolated and hard-coded, limiting interoperability, reusability, and knowledge sharing. Ontologies and knowledge graphs offer a structured and interpretable way to represent tasks, environments, and robot capabilities. Existing ontologies, such as the Socio-physical Model of Activities (SOMA) and the Descriptive Ontology for Linguistic and Cognitive Engineering (DOLCE), provide models for representing activities, spatial relationships, and reasoning structures. However, they often focus on specific domains and do not fully capture the connection between environment, action, robot's capabilities, and system-level integration. In this work, we propose the Ontology for ro\textit{BO}ts and ac\textit{T}ions (OntoBOT), which builds upon and extends existing ontologies to provide a unified representation of tasks, actions, environments, and capabilities. Our contributions are twofold: (1) we unify these core aspects into a cohesive ontology to support formal reasoning about task execution, and (2) we demonstrate its generalizability by evaluating competency questions across four distinct embodied agents — TIAGo, HSR, UR3, and Stretch — each with different capabilities, showing how OntoBOT enables context-aware reasoning and facilitates task-oriented execution, thereby promoting knowledge sharing in service robotics.
\end{abstract}

\ccsdesc[500]{Software and its engineering~Interoperability}
\ccsdesc[500]{Computer systems organization~Robotics}
\ccsdesc[500]{Information systems~Ontologies}

\keywords{Personal Service Robots, Ontology, Interoperability, System Integration}


\maketitle

\section{Introduction}
Personal service robots are increasingly deployed to assist in domestic environments, particularly in support of older adults and individuals requiring physical assistance~\cite{DBLP:journals/corr/abs-2304-14944,DBLP:journals/arcras/NanavatiRC24,holland2021service,roy2000towards,sorensen2024care}. The tasks that service robots typically perform range from cleaning and tidying to meal preparation and care-giving, often within open-ended, dynamic, and unstructured environments. Effective operation in these settings requires a high-level understanding of the context in which tasks are performed, the specific actions involved, and the robot's own physical and functional capabilities \cite{paulius2019survey, bajd2010robotics}. This process depends on the integration of hardware components (e.g. sensors, actuators) with software systems capable of representing and reasoning about tasks, environments, and the robot's own capabilities. Middleware platforms like the Robot Operating System (ROS)\footnote{\url{https://www.ros.org/}} \cite{quigley2009ros} offer open-source tools and libraries to help connect low-level hardware with higher-level functionalities. However, real-world deployments are often tightly coupled to isolated and platform-dependent solutions \cite{axelsson2015systematic,garcia2023software,wang2024survey}, with hard-coded instructions designed for specific hardware and software stacks \cite{axelsson2015systematic,garcia2023software}, limiting interoperability, knowledge sharing, cross-domain reusability and system integration.

Ontologies and Knowledge Graphs (KG) offer a promising approach for representign robotic knowledge in a structured, shareable, and machine-interpretable knowledge \cite{paulius2019survey}. Several ontologies relevant to robotics have been proposed, including efforts toward standardization such as the IEEE Robotics and Automation Society (RAS) Ontologies \cite{prestes2013towards,gonccalves2021ieee,towards_robot_task_ontology_standard}, representations like the Ontology for Collaborative Robotics and Adaptation (OCRA) \cite{ocra2022} and the Autonomous Robot Task Processing Framework (ART-ProF) \cite{artprof2024}, and foundational models such as the Descriptive Ontology for Linguistic and Cognitive Engineering (DOLCE) \cite{masolo2003descriptive,borgo2022dolce} and the Socio-physical Model of Activities (SOMA) \cite{bessler2021foundations}. These ontologies provide a formal representation for reasoning about objects, actions, spatial relationships, and activities in physical and social contexts. However, they focus on specific aspects of robotic systems, such as task planning or perception, and do not capture the full interplay between a robot's capabilities, the actions required to complete a task, and the environment in which those tasks occur. This creates a gap when reasoning about whether a robot is suited to perform a given task in a particular context, especially in heterogeneous, dynamic and collaborative settings. For example, domestic environments such as kitchens often involve recurring tasks and spatial configurations. While the robots deployed in these settings may vary in hardware and capabilities, the structure of the environment and the nature of the tasks often remain comparable. Existing ontologies can partially represent these individual elements -- such as actions, objects, or spatial relations -- but they lack an integrated way to connect between each other, and to robot-specific capabilities. 

In this work, we introduce the \textbf{Ontology for ro\textit{BO}ts and ac\textit{T}ions} (OntoBOT)\footnote{\url{https://w3id.org/onto-bot}}, a unified model designed to formally represents the interconnected elements of robotic task execution, namely: the robot and its capabilities, the task it has to perform, and the environment in which it operates. Through OntoBOT, we achieved two key contributions: unifying these core aspects into a cohesive semantic framework, and enabling reasoning about task execution across embodied agents, as demonstrated through the evaluation of competency questions involving four different robots — TIAGo, HSR, UR3, and Stretch — with varying capabilities. All materials are publicly available on GitHub\footnote{\url{https://github.com/kai-vu/OntoBOT}}, including a Jupyter notebook specifically developed for reproducing the competency question evaluation\footnote{\url{https://github.com/kai-vu/OntoBOT/blob/main/case-study/cqs.ipynb}}.

\section{Background}

We review here two key areas: the challenges and requirements of personal service robots, and the role of ontologies in supporting symbolic knowledge and interoperability in robotics. 

\subsection{Personal Service Robots}
\label{sec:personal-service-robots}
Personal service robots are autonomous or semi-autonomous systems designed to assist humans with tasks in domestic, care-giving, or socially-oriented settings. While there is no universally agreed-upon definition, the International Federation of Robotics defines them as systems that ``perform useful tasks for humans or equipment, excluding industrial automation applications''\footnote{\url{https://ifr.org/img/office/Service_Robots_2016_Chapter_1_2.pdf}}, with classification into industrial or service categories depending on intended use. Others emphasize the integration of artificial intelligence, natural language communication, and human interaction capabilities to enable adaptation in dynamic environments \cite{wirtz2018brave,kopacek2016development,belanche2020service}.

These robots are increasingly deployed in human-centred spaces such as homes and care facilities, where tasks are varied, environments are unstructured, and user needs evolve dynamically \cite{paulius2019survey,roy2000towards,sorensen2024care}. Tasks may include navigation, manipulation, object fetching, meal preparation, and human interaction. Despite advances in perception, planning, and navigation control \cite{collins2020improving,mcleay2021replaced}, most deployed systems remain siloed. Instructions are often hard-coded, tightly coupled to specific hardware and software stacks, limiting cross-domain reusability and hindering system integration \cite{axelsson2015systematic,garcia2023software}. Robotic middleware platforms such as the Robot Operating System (ROS)\footnote{\url{https://www.ros.org/}}  \cite{quigley2009ros} have significantly improved architectural modularity, enabling standard communication between components. However, middlewares primarily operates at the software infrastructure level and lacks a semantic layer - i.e. how tasks, actions, capabilities, and environments are represented and reasoned about symbolically. To support flexible, explainable, and interoperable behaviour, there is a need for shared, machine-readable representations of robot knowledge and physical capabilities \cite{wang2024survey,HAIDEGGER20131215}. This work address this gap by proposing a unified ontology that captures task structure, environmental context, and robot capabilities.

\subsection{Ontologies for Robotics}

Ontologies and Knowledge Graphs (KGs) offer a formal, structured, and interpretable way to represent knowledge in robotic systems, and a range of ontologies has been developed for this purpose. For example, ~\cite{phdthesis} presents an ontology linking sensor data with context-aware reasoning for indoor robots, while ~\cite{10.1145/3445034.3460506} models social robot services, connecting low-level perception to high-level decision-making. Similarly, ARTProF~\cite{10.3389/fnbot.2024.1401075} integrates symbolic planning and action selection. Despite their utility, many such models are tightly coupled to specific domains or platforms, limiting reusability.

More modular approaches include the Task-Skill-Resource (TSR) model \cite{KR2021-73}, which links symbolic tasks to required skills and resources, and planning integrations like \cite{ir.2021.10}. However, these often lack grounding in physical robot capabilities or environmental affordances. Architecture-level frameworks have also been proposed:  ~\cite{skarzynski2017somrsmultirobotarchitecturebased} uses service-oriented design to support heterogeneous robot collaboration, and~\cite{zander2016modeldrivenengineeringapproachros} applies semantic models to ROS component development. Further, ~\cite{10.1145/3445034.3460506} proposes a reference architecture for evaluating interoperability using scenario-based metrics.

Several works have emerged employing foundational ontologies to model robotic knowledge. For example, the Descriptive Ontology for Linguistic and Cognitive Engineering (DOLCE) \cite{masolo2003descriptive,borgo2022dolce}, is widely used as an upper ontology for modeling entities, events, and participation roles. It is the base of frameworks like the Socio-physical Ontology Model of Activities (SOMA) \cite{bessler2021foundations}, which models embodiment, affordances, and spatial reasoning in robot-world interactions. ROS.owl~\cite{tiddi2017ontology} describes execution-level semantics of ROS systems but lacks symbolic or environmental context. Ontologies such as the Ontology for Provenance and Plans (PPlan)~\footnote{\url{https://vocab.linkeddata.es/p-plan/index.html}} and the Procedural Knowledge Ontology (PKO)~\cite{carriero2025procedural} model procedural and plan-based knowledge but are often disconnected from execution or perception layers. While each ontology addresses important facets, they are typically developed in isolation with limited alignment, failing to capture the robotic behavior pipeline: from user intent and symbolic planning, through environmental interaction, to platform-specific integration.

To address this gap, we propose a unified ontology that connects symbolic actions with procedural structure, environmental context, and robot capabilities. Rather than replacing existing models, OntoBOT aligns and integrates them through shared concepts, supporting interoperable and task-oriented robotic behavior.

\section{The OntoBOT Ontology}
\label{sec:ontoBOT}

\begin{figure*}[t]
  \centering
  \includegraphics[width=0.8\textwidth]{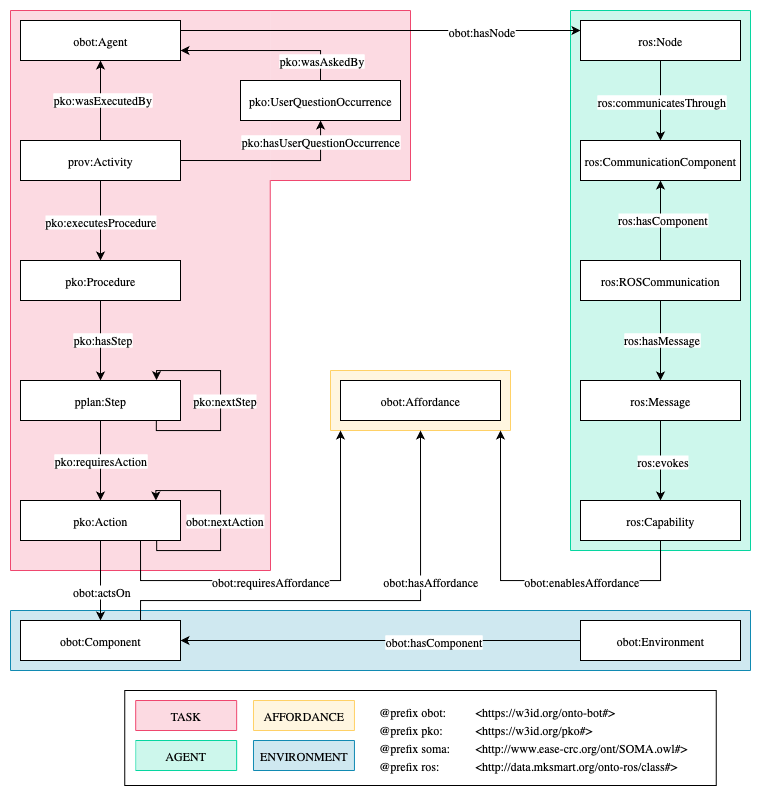}
  \caption{Overview of the OntoBOT ontology and its core concepts.}
  \label{fig:ontobot}
\end{figure*}

To support reasoning and interoperability in task-orient service robotics, we developed the \textbf{Ontology for ro\textit{BO}ts and ac\textit{T}ions} (OntoBOT)\footnote{\url{https://w3id.org/onto-bot}}, an ontology based on the Web Ontology Language (OWL) \cite{hitzler2009owl}. OntoBOT models how an embodied agent (robot) interacts with its environment, and perform tasks based on its capabilities. Figure \ref{fig:ontobot} gives an overview of OntoBOT's core concepts and their relationships. OntoBOT was developed following the Ontology 101 methodology \cite{noy2001ontology}, including scope definition, competency questions, and reuse of existing ontologies. A set of Competency Questions (CQs) was formulated in collaboration with robotic experts, and serve as both design guidelines and validation metrics (see Section~\ref{sec:case-study}). OntoBOT builds upon, integrates and extends elements from the following ontologies: DOLCE (@prefix dul: <http://www.ontologydesignpatterns.org/ont/dul/DUL.owl\#>), SOMA (@prefix soma: <http://www.ease-crc.org/ont/SOMA.owl\#>), PKO (@prefix pko: <https://w3id.org/pko\#>) and ROS.owl (@prefix ros: <http://data.mksmart.org/onto-ros/class\#>). In line with the goal of promoting interoperability and avoiding redundant modeling, OntoBOT introduces a minimal set of new elements (3 new classes and 7 new object properties). To clarify the scope of OntoBOT and the concepts it models, we introduce the following definitions: \\
{\textbf{Agent.}} Following the definitions in \ref{sec:personal-service-robots}, an agent refers to an autonomous or semi-autonomous robot that can perceive its environment, interact with humans, and perform meaningful tasks. It typically possesses capabilities such as navigation, manipulation, and object recognition. \\
\textbf{Environment.} The physical context in which the agent operates, consisting of entities such as objects, furniture, and appliances that the robot can perceive and interact with. \\
{\textbf{Task.}} A structured, goal-oriented activity composed of one or more actions. Tasks represent high-level procedures initiated by user goals, and they are decomposed into executable actions through symbolic planning. \\
{\textbf{Affordance.}} A relational concept describing the interaction between an agent and a physical component. Originating from ecological psychology~\cite{gibson2014ecological}, affordances depend not only on object properties but also on the agent's morphology and capabilities. In OntoBOT, affordances are made machine-readable and linked to executable actions for logical reasoning.


\subsection{Modeling the Agent}
\label{sec:modelingagent}
OntoBOT represents acting entities with the class \texttt{obot:Agent}, a subclass of \texttt{dul:Agent}, \texttt{prov:Agent} and \texttt{foaf:Agent}. Robotic execution is modeled via ROS.owl: each agent is linked to one or more software \texttt{ros:Node}s via \texttt{obot:hasNode}, which communicate through \texttt{ros:CommunicationComponent}s and exchange \texttt{ros:Message}s. Messages can \texttt{ros:evoke} specific \texttt{ros:Capability} instances - executable functions based on the robot's software and hardware. OntoBOT introduces the property \texttt{obot:enablesAffordance} to link capabilities and affordances, allowing reasoning over whether an agent can act upon an object. For example, a ``graspable'' object affords grasping, but this affordance is only actionable if the agent has a capability that enables it.

\subsection{Modeling the Environment}
The physical world is described by the class \texttt{obot:Environment}, representing a concrete space where agents perceive, plan, act. Environments consist of \texttt{obot:Component}s - physical entities that the robot can interact with - linked via the property \texttt{dul:hasComponent}. \texttt{obot:Environment} is defined as subclass of \texttt{dul:Place}. While the original DUL definition refers to a wide range of locations, including abstract or relational ones, OntoBOT narrows this scope by interpreting environments as physically bounded spaces, where agents can detect, navigate, and act upon components. Each component in the environment can be associated with one or more affordances using the property \texttt{obot:hasAffordance}, indicating what that object affords — for example, being graspable, holdable, or openable.

\subsection{Modeling the Affordance}
Affordances link symbolic actions, physical components, and robot capabilities. OntoBOT introduces the class \texttt{obot:Affordance}, a subclass of both \texttt{soma:Affordance} and \texttt{soma:PhysicalTask}, to model physical interactions such as grasping, holding and opening. Rather than reusing \texttt{soma:Affordance} directly, OntoBOT adopts a Gibsonian view \cite{gibson2014ecological}, treating affordances not as static object properties but as relational, context-dependent possibilities for action. While existing representations, including SOMA's, often treat affordances as statitic, object-centered labels -- for example, a drawer is \textit{openable} -- OntoBOT explicitly connects them to the component, the symbolic action, and the agent capable of enacting them. This richer representation supports reasoning about which agents can perform which tasks given an environment, grounded in both task semantics and physical capability. 

\subsection{Modeling the Task}

OntoBOT structures robot behavior using procedural models, drawing from PKO and PPlan. Task execution is formalized as a \texttt{prov:Activity}, triggered by a user query or internal goal, and it is linked to the executing agent via \texttt{prov:wasAssociatedWith}. Each activity \texttt{pko:executesProcedure}, which describes the planned method for achieving a goal. A procedure consists of multiple \texttt{pplan:Step}s, ordered through the property \texttt{pko:nextStep}, and defined via \texttt{pko:hasStep}. Each step  \texttt{pko:requiresAction}s, as instances of \texttt{pko:Action}, which are atomic operation expected of the agent, ordered though \texttt{obot:nextAction}. Each action targets a component using the property \texttt{obot:actsOn}, and is semantically linked to one or more affordances via \texttt{obot:requiresAffordance}. This ensures that an action is only feasible if the environment provides the necessary interactions and the agent has the capability to realize it. For instance, the task ``retrieve a spoon from the drawer'', consists of actions like opening the drawer and grasping the spoon. Each action is linked to affordances (e.g., \textit{openable}, \textit{graspable}) and constrained by the agent's physical capabilities. If an agent lacks the required capabilities, OntoBOT supports reasoning that the task is not executable, ensuring procedural knowledge is grounded in environmental context and agent embodiment.






\section{Case Study}
\label{sec:case-study}
To evaluate OntoBOT, we investigate a case study involving two sequential activities and four different robotic platforms with varying capabilities. The scenario has been developed in collaboration with project partners and domain experts in robotics, and it aligns with the European Robotic League Coopetitions\footnote{\url{https://www.eurobin-project.eu/index.php/competitions/coopetitions}}, which serve as a practical benchmark for assessing real-world robotic systems in domestic environments.  The case study is set in a household kitchen and includes the activities of ``preparing breakfast'' and ``reorganising the kitchen'' after the meal. These scenarios were chosen to illustrate how OntoBOT models the interplay between environments, robot capabilities, and task requirements. Alongside the activities, we also include knowledge about four different robots — TIAGo\footnote{\url{https://pal-robotics.com/robot/tiago/}}, HSR\footnote{\url{http://www.hsrobotics.net/products-view-51.html}}, UR3\footnote{\url{https://www.universal-robots.com/products/ur3e/}}, and Stretch\footnote{\url{https://bostondynamics.com/products/stretch/}} — each with different morphology and capabilities. This allows us to evaluate how OntoBOT generalizes across platforms and supports reasoning about capability-task requirement. The setting of the Coopetition also informed the formulation of the competency questions (CQs), as the types of knowledge extracted can directly support informed decision-making in competition settings. Further details on the activities and robots are provided in Sections~\ref{sec:activities} and ~\ref{sec:robots}.

The formulated CQs as validation measure are as follows:

\begin{enumerate}
    \item What objects and their associated affordances are involved in the activity ``prepare breakfast''?
    \item What is the required sequence of actions to complete the ``prepare breakfast'' activity?
    \item What capabilities are required for a robot to perform a given activity?
    \item Which robot is capable of executing all actions required for a given activity?
    \item Given its capabilities, can robot X execute both the ``prepare breakfast'' and ``reorganise the kitchen'' activities?
    \item If not, which capabilities does Robot X lack, and which steps in the activity are unachievable as a result?
\end{enumerate}

\subsection{Activities}
\label{sec:activities}
Both activities take place in a household kitchen and involve a controlled set of objects and actions representative of common domestic tasks. The first - ``prepare breakfast'' - includes making cereal with milk and serving orange juice, requiring retrieval of items like a bowl, spoon, glass, cereal box, and bottles. The second - ``reorganise the kitchen'' - focuses on clean-up and storage, such as returning items to their places or placing them in the dishwasher.

\subsection{Robots}
\label{sec:robots}
Our case study considers four robots with different morphologies and capabilities: TIAGo, HSR, UR3 and Stretch. While each platform supports a wide range of functionalities, we focus on a selected subset relevant to task execution of the activities: grasping, holding, and placing objects, opening and closing drawers or cupboard, and pouring liquids. 
{\textbf{TIAGo}} features an arm with a gripper that enables reliable grasping, holding, placing objects, and pouring liquids\footnote{\url{https://pal-robotics.com/blog/tiago-robot-iros-mobile-manipulation-hackathon}}, as well as opening and closing drawers \cite{magyar2019guided}.
{\textbf{HSR}} is equipped with a gripper and vacuum pad, and is capable of grasping, holding, placing objects\footnote{\url{https://www.toyota-global.com/innovation/partner_robot/robot/file/HSR_EN.pdf}} and opening drawers \cite{yamamoto2019development}, but to the best of our knowledge, it has not been reported to be able to pour liquids. 
{\textbf{UR3}} is a collaborative robotic arm capable of grasping, holding, placing and pouring \cite{proesmans2025instrumentation}. However, while force-controlled opening/closing capabilities have been explored \cite{perez2020force}, it is not supported in the standard configuration and is thus excluded from its available capabilities in this research. 
{\textbf{Stretch}} is designed for mobile manipulation, but particularly in logistic environments, and is able to grasp, hold and place objects \footnote{\url{https://bostondynamics.com/blog/introducing-multipick-for-automated-unloading/}}. However, we do not have evidences that Stretch currently support opening/closing and pouring actions. 

\subsection{Evaluation}
As evaluation, we instantiated KGs representing the activities and the robots, structured according OntoBOT. Encoded in TTL format, the activity KG captured the environment, its objects and affordances. The robot KG details the functional capabilities of each robot. We then assessed OntoBOT by translating the CQs into SPARQL queries and executing them against the KGs using the Python RDFLib\footnote{\url{https://rdflib.readthedocs.io/en/stable/}} library.

\section{Results}
This section presents the results of the CQs executed as SPARQL queries over the KGs constructed following OntoBOT: one modeling the activities (421 triples) and one modeling the robots (50 triples). Due to space constraints, we provide excerpts of the query results here. The full implementation, including all queries and results, is available on GitHub\footnote{\url{https://github.com/kai-vu/OntoBOT}}, along with a Jupyter notebook\footnote{\url{https://github.com/kai-vu/OntoBOT/blob/main/case-study/cqs.ipynb}} that allows for replication and further exploration of the use case. The SPARQL queries reuse the prefixes defines in Section \ref{sec:ontoBOT}, and instances for the use case are defined using @prefix : <https://example.org/>.

\paragraph{\textbf{CQ1: What objects and their associated affordances are involved in the activity ``prepare breakfast''?}} 
The first CQ aims to identify the physical objects relevant to an activity and the types of affordances they enable. Understanding this relationship is essential for assigning appropriate tasks to robots, which rely on affordance-level information to determine how they can interact with objects in the environment. Listing \ref{lst:cq1} retrieves all distinct object-affordance pairs by following the \texttt{obot:actsOn} and \texttt{obot:requiresAffordance} properties from actions to their associated objects and affordances. Table \ref{tab:cq1} shows a subset of the results for CQ1. For example, the object \texttt{:orangeJuice} is associated with several affordances such as \texttt{soma:Grasping}, \texttt{soma:Holding}, \texttt{soma:Placing} and \texttt{soma:Pouring}.

\vspace{-10pt}

\begin{figure}[h]
\begin{center}
\begin{minipage}{0.9\linewidth}
\begin{lstlisting}[caption={SPARQL query for CQ1\label{lst:cq1}.}]
SELECT DISTINCT ?object ?affordance
WHERE {
    ?activity a prov:Activity ;
        rdfs:label "Prepare breakfast" ;
        pko:executesProcedure ?procedure .
    ?procedure pko:hasStep ?step .
    ?step pko:requiresAction ?action .
    ?action obot:actsOn ?object ;
        obot:requiresAffordance ?affordance . }
\end{lstlisting}
\end{minipage}
\end{center}
\end{figure}

\vspace{-20pt}

\begin{table}[h]
\centering
\captionsetup{belowskip=0pt}
\caption{Results for CQ1.}
\label{tab:cq1}
\vspace{-11pt}
\rowcolors{2}{gray!10}{white}
{\scriptsize
\begin{tabular}{|p{1.5cm}|p{5.7cm}|}
\hline
\rowcolor{gray!30}
\multicolumn{1}{|c|}{Object} & \multicolumn{1}{c|}{Affordance} \\
\hline
:drawer & soma:Opening, soma:Closing \\
:bowl & soma:Grasping, soma:Holding, soma:Placing \\
:orangeJuice & soma:Grasping, soma:Holding, soma:Placing, soma:Pouring \\
\hline
\end{tabular}
}
\end{table}

\vspace{-10pt}
\paragraph{\textbf{CQ2: What is the required sequence of actions to complete the ``prepare breakfast'' activity?}} 
This CQ focuses on extracting the sequential procedural structure of a given activity. Answering this CQ helps determine how high-level tasks can be described into ordered, atomic actions, which can facilitate task planning and execution in robotic systems. The associated SPARQL query - Listing \ref{lst:cq2} - extracts the activity's procedure, its steps, and the actions associated with each step. In this example, we filter on the ``prepare breakfast'' activity. The result of the SPARQL query shows how the activity is decomposed into a sequence of high-level procedures, namely \texttt{Retrieve tableware}, \texttt{Retrieve food}, and \texttt{Serve food}. Within each procedure, the steps (e.g. \texttt{Serve milk}, \texttt{Serve cereal}) are defined in a sequential order using the \texttt{pko:nextStep} property. Similarly, each step is composed of more fine-grained actions (\texttt{pko:Action}) that are also ordered using the \texttt{obot:nextAction} property. The excerpt in Table \ref{tab:cq2} show results for the \texttt{Serve food} procedure, and demonstrates how it is composed of steps such as \texttt{Serve milk}, \texttt{Serve orange juice}, and \texttt{Serve cereal}. Each of these steps is linked to a sequence of actions representing real-world robotic tasks, such as \texttt{Grasp the milk}, \texttt{Pour the milk into the bowl}, and \texttt{Put milk down}.


\begin{figure}[h]
\begin{center}
\begin{minipage}{0.9\linewidth}
\begin{lstlisting}[caption={SPARQL query for CQ2\label{lst:cq2}.}]
SELECT DISTINCT ?activity ?procedureLabel ?stepLabel ?actionLabel
WHERE {
    ?activity a prov:Activity ;
        rdfs:label "Prepare breakfast" ;
        pko:executesProcedure ?procedure .
    ?procedure rdfs:label ?procedureLabel ;
        pko:hasStep ?step .
    ?step rdfs:label ?stepLabel ;
        pko:requiresAction ?action .
    ?action rdfs:label ?actionLabel . }
\end{lstlisting}
\end{minipage}
\end{center}
\end{figure}


\begin{table}[h]
\centering
\captionsetup{belowskip=0pt}
\caption{Results for CQ2.}
\label{tab:cq2}
\vspace{-11pt}
{\scriptsize
\begin{tabular}{|p{1.5cm}|p{2.2cm}|p{3cm}|}
\hline
\rowcolor{gray!30}
\multicolumn{1}{|c|}{Procedure} & \multicolumn{1}{c|}{Step} & \multicolumn{1}{c|}{Action} \\
\hline
\rowcolor{white}
\multirow[c]{3}{*}{\makecell[c]{Serve food}} & Serve milk & Grasp the milk \newline Pour milk into the bowl \newline Put milk down \\
\cellcolor{white} & \cellcolor{gray!10}Serve orange juice & \cellcolor{gray!10}Grasp the orange juice \newline \cellcolor{gray!10}Pour the orange juice \newline \cellcolor{gray!10}Put orange juice down \\
\cellcolor{white} & \cellcolor{white}Serve cereal & \cellcolor{white}Grasp the cereal box \newline \cellcolor{white}Pour cereal into the bowl \newline \cellcolor{white}Put cereal box down \\
\hline
\end{tabular}
}
\end{table}

\vspace{-5pt}
\paragraph{\textbf{CQ3: What capabilities are required for a robot to perform a given activity?}} 
The SPARQL query associated with CQ3 (Listing \ref{lst:cq3}) is designed to retrieve all the affordances required to perform the actions that constitute a given activity. The query traverses the ontology structure by first identifying an activity and its associated procedures and steps, then the actions required at each step, and finally retrieving the affordances required to perform each action. These affordances are linked to the actions via the \texttt{obot:requiresAffordance} property. The rationale behind this query is to determine the set of capabilities a robot must possess in order to complete a given activity. This information can then be used to assess whether a specific robot - based on its known capabilities - can execute the task or not (as further explored in CQ 4 and CQ 5). The results in Table \ref{tab:cq3} show that the activity \texttt{Prepare breakfast} requires a broader range of affordances, including \texttt{soma:Grasping}, \texttt{soma:Holding}, \texttt{soma:Placing}, \texttt{soma:Pouring}, \texttt{soma:Opening} and \texttt{soma:Closing}. In contrast, the activity \texttt{Reorganise the kitchen} involves a slightly reduced set of affordances, lacking \texttt{soma:Pouring}, which aligns with the nature of the activity being more focused on moving and storing items rather than manipulating or serving food. This capability breakdown is essential for enabling reasoning over robot-task compatibility and can facilitate evaluating whether a robot is equipped to handle the demands of a given task and environment. 

\vspace{-10pt}

\begin{figure}[h]
\begin{center}
\begin{minipage}{0.9\linewidth}
\begin{lstlisting}[caption={SPARQL query for CQ3.\label{lst:cq3}}]
SELECT DISTINCT ?activityLabel ?affordance
WHERE {
    ?activity pko:executesProcedure ?procedure ;
        rdfs:label ?activityLabel .
    ?procedure pko:hasStep ?step .
    ?step pko:requiresAction ?action .
    ?action obot:requiresAffordance ?affordance . }
\end{lstlisting}
\end{minipage}
\end{center}
\end{figure}

\vspace{-30pt}

\begin{table}[h]
\centering
\captionsetup{belowskip=0pt}
\caption{Results for CQ3.}
\label{tab:cq3}
\vspace{-11pt}
\rowcolors{2}{gray!10}{white}
{\scriptsize
\begin{tabular}{|p{2cm}|p{5.1cm}|}
\hline
\rowcolor{gray!30}
\multicolumn{1}{|c|}{Activity} & \multicolumn{1}{c|}{Affordance} \\
\hline
Prepare breakfast & soma:Grasping, soma:Holding, soma:Placing, soma:Pouring \newline soma:Opening, soma:Closing \\
Reorganise the kitchen & soma:Grasping, soma:Holding, soma:Placing \newline soma:Opening, soma:Closing \\
\hline
\end{tabular}
}
\end{table}

\vspace{-5pt}
\paragraph{\textbf{CQ4: Which robot is capable of executing all actions required for a given activity?}} 
This competency question aims to determine which robot possess the necessary capabilities to carry out all actions involved in a given activity. To answer this, we split the query into two parts due to technical constraints, namely the SPARQL \texttt{HAVING} clause used to enforce completeness (e.g. a robot must match \textit{all} required affordances) that was not supported by the SPARQLWrapper Python library used in the Jupyter notebook. As a result, we execute two separate SPARQL queries. The first retrieves all affordances required by the actions in the target activity. The second retrieves all affordances each robot can enable, based on their functional capabilities. these affordances are evoked ROS messages and linked to physical capabilities using \texttt{obot:enablesAffordance}. Post-query processing was done in Python, where we compared the capabilities of each robot with the required affordances. A robot is considered capable of performing the activity only if it has the ability to perform \textit{all} affordances required. The results revealed that for the activity ``\texttt{Prepare breakfast}'', only the TIAGo robot met all the required affordances. In contrast, for ``\texttt{Reorganise the kitchen}'', both TIAGo and HSR satisfied the affordance requirements. This results reflects the functional demands of the first activity, which involves more fine-grained manipulation tasks such as pouring, in addition to grasping and placing. 


\begin{figure}[h]
\begin{center}
\begin{minipage}{0.9\linewidth}
\begin{lstlisting}[caption={SPARQL query for CQ4.\label{lst:cq4}}]
SELECT DISTINCT ?affordance
WHERE {
    ?activity pko:executesProcedure ?procedure ;
        rdfs:label "Reorganise the kitchen" .
    ?procedure pko:hasStep ?step .
    ?step pko:requiresAction ?action .
    ?action obot:requiresAffordance ?affordance . }
-----------------------------------------------
SELECT DISTINCT ?robotLabel ?affordance
WHERE {
    ?robot a obot:Agent ;
        rdfs:label ?robotLabel ;
        obot:hasNode ?node .
    ?node ros:communicatesThrough ?commComponent .
    ?comm a ros:ROSCommunication ;
        ros:hasComponent ?commComponent ;
        ros:hasMessage ?msg .
    ?msg ros:evokes ?capability .
    ?capability obot:enablesAffordance ?affordance . }
\end{lstlisting}
\end{minipage}
\end{center}
\end{figure}

\paragraph{\textbf{CQ5: Given its capabilities, can robot X execute both the ``prepare breakfast'' and ``reorganise the kitchen'' activities?}} 
Building on CQ 4, this question evaluates whether a specific robot has the combined set of capabilities needed to perform all activities. As with CQ 4, the answer was derived using two separate SPARQL queries: one to collect all affordances required across both activities and another to list the affordances enables by each robot. The comparison was again implemented in Python. The result confirmed that TIAGo is the only robot capable of performing both activities, and the outcome is consistent with the results from CQ4. 


\begin{figure}[h]
\begin{center}
\begin{minipage}{0.9\linewidth}
\begin{lstlisting}[caption={SPARQL query for CQ5.\label{lst:cq5}}]
SELECT DISTINCT ?reqAff
WHERE {
    ?activity a prov:Activity ;
        pko:executesProcedure ?procedure .
    ?procedure pko:hasStep ?step .
    ?step pko:requiresAction ?action .
    ?action obot:requiresAffordance ?reqAff . }
-----------------------------------------------
SELECT DISTINCT ?robotLabel ?affordance
WHERE {
    ?robot a obot:Agent ;
        rdfs:label ?robotLabel ;
        obot:hasNode ?node .
    ?node ros:communicatesThrough ?commComponent .
    ?comm a ros:ROSCommunication ;
        ros:hasComponent ?commComponent ;
        ros:hasMessage ?msg .
    ?msg ros:evokes ?capability .
    ?capability obot:enablesAffordance ?affordance . }
\end{lstlisting}
\end{minipage}
\end{center}
\end{figure}

\paragraph{\textbf{CQ6: If not, which capabilities does Robot X lack, and which steps in the activity are unachievable as a result?}} 
This question investigates which specific affordances are required by individual steps in both activities cannot be met by the capabilities of each robot, thereby identifying the steps that are unachievable. To answer this, two SPARQL queries were also employed: the first extracts all the affordances required by each step in the activities, while the second retrieves the affordances enables by each robot based on their capabilities. A comparison of these sets highlights the missing affordances for each robot and the consequent steps they cannot perform. The results, summarized in Table \ref{tab:cq6}, indicate that HSR cannot perform the ``serving food'' step as it lacks the capability to meet the affordance \texttt{soma:Pouring}. UR3 lacks capabilities for \texttt{soma:Opening} and \texttt{soma:Closing}, which are necessary for accessing cupboards and the dishwasher, but it can serve food since that step does not require those affordances. Stretch, despite having grasping, opening, and closing capabilities, cannot complete any of the steps fully due to its inability to meet the affordance \texttt{soma:Holding}. TIAGo, instead, has the capabilities to perform all steps in both activities. 


\begin{table}[h!]
\centering
\caption{Results for CQ6.}
\label{tab:cq6}
\vspace{-11pt}
{\scriptsize 
\begin{tabular}{|p{1.3cm}|p{1.7cm}|C{0.6cm}|C{0.6cm}|C{0.6cm}|C{0.6cm}|}
\hline
\rowcolor{gray!30}
\multicolumn{1}{|c|}{} & \multicolumn{1}{c|}{} & TIAGo & HSR & UR3 & Stretch \\
\hline
\multirow[c]{3}{*}{\makecell[c]{Prepare \\ breakfast}} 
    & Retrieve tableware & \cmark & \cmark & \xmark & \xmark \\
    & \cellcolor{gray!10}Retrieve food & \cellcolor{gray!10}\cmark & \cellcolor{gray!10}\cmark & \cellcolor{gray!10}\xmark & \cellcolor{gray!10}\xmark \\
    & Serve food  & \cmark & \xmark & \cmark & \xmark \\
\hline
\multirow[c]{2}{*}{\makecell[c]{Reorganise \\ the kitchen}} 
    & \cellcolor{gray!10}Put away food & \cellcolor{gray!10}\cmark & \cellcolor{gray!10}\cmark & \cellcolor{gray!10}\xmark & \cellcolor{gray!10}\xmark \\
    & Load dishwasher & \cmark & \cmark & \xmark & \xmark \\
\hline
\end{tabular}
}
\end{table}

\begin{figure}[h!]
\begin{center}
\begin{minipage}{0.9\linewidth}
\begin{lstlisting}[caption={SPARQL query for CQ6.\label{lst:cq6}}]
SELECT DISTINCT ?activityLabel ?stepLabel ?affordance
WHERE {
    ?activity a prov:Activity ;
        rdfs:label ?activityLabel ;
        pko:hasUserQuestionOccurrence ?uqo ;
        pko:executesProcedure ?procedure .
  ?procedure pko:hasStep ?step .
  ?step rdfs:label ?stepLabel ;
        pko:requiresAction ?action .
  ?action obot:requiresAffordance ?affordance . }
-----------------------------------------------
SELECT DISTINCT ?robotLabel ?affordance
WHERE {
  ?robot a obot:Agent ;
         rdfs:label ?robotLabel ;
         obot:hasNode ?node .
  ?node ros:communicatesThrough ?commComponent .
  ?comm a ros:ROSCommunication ;
        ros:hasComponent ?commComponent ;
        ros:hasMessage ?msg .
  ?msg ros:evokes ?capability .
  ?capability obot:enablesAffordance ?affordance . }
\end{lstlisting}
\vspace{-1.4\baselineskip}
\end{minipage}
\end{center}
\end{figure}


\section{Conclusion}
This work addresses the challenge of developing a unified model that formally represents the interconnected elements of robotic task execution, specifically: the robot and its capabilities, the task to be performed, and the environment in which these tasks take place. Through OntoBOT, we achieved two key contributions: unifying these core aspects into a cohesive semantic framework, and enabling consistent reasoning about task execution across diverse embodied agents, as demonstrated through CQs evaluation involving multiple robots with varying capabilities.

OntoBOT provides a structured, semantic representation of robotic systems, actions, and environmental interactions. By encoding symbolic tasks as procedures composed of steps - each requiring specific actions and, by extensions, specific affordances - the model provides a formalization of task requirements. At the same time, robot capabilities are modeled in terms of the affordances they enable, allowing direct comparison between what is required and what is possibly for a given agent. Our framework treats symbolic actions as executable only when both the environment affords the necessary interaction and the robot possesses the required capabilities. This affordance-centric reasoning is central to our framework, offering a scalable approach for determining action feasibility and explaining when tasks are only partially achievable. The case study and its relative competency questions addressed in this work illustrate the practical implications of OntoBOT. It supports interpretable, knowledge-based decision-making by identifying relevant objects and affordances (CQ1), extracting task structure (CQ2), and determining required capabilities (CQ3), as well as checking task feasibility (CQs 4 and 5) and identifying missing capabilities that prevent step execution (CQ6).

Future directions will focus on automation. Recent works in neurosymbolic approaches show how KGs can be generated using multimodal language models coupled with ontologies, starting from unstructured sensory data such as images~\cite{martorana2025bridging}. Moreover, we aim to extend OntoBOT to include richer spatial and contextual descriptions of the environment. For example, modeling the state of the environment before and after each action would allow for reasoning not only about whether an action is possible, but also whether a task has been successfully completed. Tracking object locations and status over time could enable robots to verify actions completion, and plan follow-up actions, a key aspect in autonomy. 

In summary, this work demonstrates how a unified ontological framework grounded in affordances and capabilities can support interpretable reasoning about robot-task-environment interactions. Future extensions with neurosymbolic knowledge generation and dynamic reasoning can further facilitate agents' ability to support autonomous decision-making in real-world settings. 


\section*{Acknowledgements}
This work is funded by the Technology Exchange Programme of the Horizon Europe euROBIN project (grant agreement No 101070596). Further, we acknowledge that ChatGPT was utilized to generate and debug part of the Python and \LaTeX code used in this work. 

\bibliographystyle{ACM-Reference-Format}
\bibliography{main}

\end{document}